\documentclass{article}
\usepackage{spconf,amsmath} 
\usepackage{optidef}
\usepackage{amssymb}
\usepackage{booktabs}       
\usepackage{adjustbox}

\usepackage{subcaption}

\newtheorem{definition}{Definition}[section]

\usepackage{xcolor}
\usepackage[linesnumbered,ruled,vlined]{algorithm2e}
\usepackage{multirow}
\usepackage{multirow}
\usepackage[hidelinks]{hyperref}
\usepackage{graphicx,epstopdf}

\SetKwInput{KwInput}{Input}                
\SetKwInput{KwOutput}{Output}              
\newcommand{\h}[1]{\mathbf{#1}}

\newcommand{\fft}{{\textnormal{fft}\,}}
\newcommand{\ifft}{{\textnormal{ifft}\,}}
\newcommand{\fro}{{\textnormal{F}\,}}

\title{Non-convex approaches for low-rank tensor completion\\ under tubal sampling
}
\name{$\textnormal{Zheng Tan}^{1}$, $\textnormal{Longxiu Huang}^{2}$, $\textnormal{HanQin Cai}^{3}$,  $\textnormal{Yifei Lou}^{4}$\thanks{The work was partially supported by   AMS Simons Travel Grant,  NSF CAREER grant, NSF DMS 1846690, and NSF DMS 2304489.} }
\address{
$^1$University of California, Los Angeles, $^2$Michigan State University, \\ $^3$University of Central Florida, $^4$University of Texas at Dallas
}

\begin{document}
%
\maketitle
\begin{abstract}
Tensor completion is an important problem in modern data analysis. In this work, we investigate a specific sampling strategy, referred to as tubal sampling. We  propose two novel  non-convex tensor completion frameworks that are easy to implement, named tensor $L_1$-$L_2$ (TL12) and tensor completion via CUR (TCCUR). We test the efficiency of both methods on synthetic data and a  color image inpainting problem. Empirical results reveal a trade-off between  the accuracy and time efficiency of these two methods in a low sampling ratio. Each of them outperforms some classical completion methods in at least one aspect. 
\end{abstract}
%
\section{Introduction}
\label{sec:intro}
Tensor, a multidimensional generalization of matrix, is a useful data structure that is arisen in various fields such as seismic imaging~\cite{Kilmer2015,popa2021improved}, image or video  processing \cite{zhou2017tensor,lu2019tensor,cai2021mode}, and recommendation system \cite{zheng2016topic, song2017based}.
This paper considers a tensor completion problem in which the measured data has missing 
entries. It is an ill-posed problem, thus requiring additional information to be imposed as a regularization. We focus on a low-rank structure of the desired tensor. 

The rank of a matrix is the number of nonzero singular values. As the rank minimization is NP-hard, a popular choice is to minimize the sum of its singular values, which is called \emph{nuclear norm} \cite{Candes2009}. By regarding nonnegative singular values as a vector, the sum of singular values is equivalent to the $L_1$ norm of this vector. With recent advances in tensor algebra~\cite{Kilmer2011}, tensor nuclear norm (TNN)~\cite{6909886} was proposed to enforce the low-rankness of a tensor.  Studies~\cite{yinLHX14,louYHX14} have demonstrated that a nonconvex $L_1$-$L_2$ model gives better identification of nonzero elements compared to the convex $L_1$ norm. Motivated by this empirical observation, we propose  a novel tensor $L_1$-$L_2$ model (TL12) for low-rank tensor completion.  Recently, nonconex models for tensor recovery have been investigated in \cite{wang2021generalized}, which does not include TL12. 


Other than seeking proper regularizations, one can rely on matrix decomposition techniques to enforce the low rankness. For example, CUR decomposition \cite{HH2020,drineas2008relative,chiu2013sublinear,HH_Perturbation2019} factorizes a matrix into the product of three smaller matrices  compared to the original size by {taking its column and row  subsets  to form the left and right matrices and } enforcing the middle matrix as a low-rank matrix. Recently, a tensor CUR (t-CUR) decomposition was proposed in \cite{wang2017missing,chen2022tensor,hamm2023generalized}. 
 In this work, we focus on a specific sampling strategy for 3-mode tensors (a tensor has three dimensions) that either takes all the samples along the third dimension or not at all, which is referred to as tubal sampling. With this sampling scheme, tensor completion can be reduced to a set of matrix completion problems. We thus
 propose an efficient Tensor Completion method via CUR (TCCUR) by adapting  a recent matrix completion approach \cite{cai2022matrix} to tensor completion.

 We conduct experiments to compare the proposed nonconvex methods with classical convex models on synthetic data and a color image inpainting problem. We observe that regularization-based tensor completion methods yield higher accuracy but at a cost of computational time compared to  decomposition methods, especially under the regime of low sampling ratios.
  The main contributions of this work are threefold:
\begin{enumerate}
    \itemsep0em 
    \item We propose the TL12 regularization to promote low-rankness for tensor completion. 
    \item  We  propose an accelerated  tensor low-rank decomposition method, referred to TCCUR.
    \item We compare these two nonconvex approaches (regularization and decomposition) for empirical guidance in real applications.
\end{enumerate}

 


\section{Notation and preliminary} \label{sec:background}
Throughout this paper,  we denote scalars by lowercase letters, vectors by bold letters, matrices by uppercase letters, and tensors by calligraphic uppercase letters. The set of the first $n$ natural numbers is denoted by $[n]:=\{1,\cdots,n\}$.  We reserve $I$, $J$, $\mathcal{I}$, and $\mathcal{J}$ as  index sets. We use $A^\dagger$ to denote the Moore--Penrose pseudo-inverse of a matrix $A$.

Given a 3-mode tensor $\mathcal{A}\in\mathbb{R}^{n_1\times n_2\times n_3}$, its $i$-th frontal slice is a matrix,  denoted by $[\mathcal{A}]_{:,:,i}$, and we refer $[\mathcal{A}]_{i,j,:}$ as 
its $(i,j)$-th tube. We use
\begin{align*}
    \hat{\mathcal{A}}=\fft(\mathcal{A},[~],3) \quad \mbox{and}\quad 
    \mathcal{A}=\ifft(\hat{\mathcal{A}},[~],3),
\end{align*}
to denote the Fourier transform and the inverse Fourier transform of $\mathcal{A}$ and $\hat{\mathcal{A}}$ along the third dimension, respectively. In what follows, we provide some important tensor definitions  \cite{Kilmer2011,chen2022tensor,zhang2016exact} that are relevant to  this work. 
\begin{definition}[t-product]Given  $\mathcal{A}\in\mathbb{R}^{n_1\times n_2\times n_3}$ and $\mathcal{B}\in\mathbb{R}^{n_2\times \ell\times n_3}$, the t-product $\mathcal{C}=\mathcal{A}*\mathcal{B}$ is an $n_1\times\ell \times n_3$ tensor and its $(i,j)$-th tube is given by
\[[\mathcal{C}]_{i,j,:}=\sum_{k=1}^{n_2}[\mathcal{A}]_{i,k,:}*[\mathcal{B}]_{k,j,:},
\]
where $*$ denotes the circular convolution between two tubes (vectors) of the same length.
\end{definition}

With t-product, one can extend the matrix pseudo-inverse into tensor, i.e., $\mathcal A^\dagger$ is the pseudo-inverse of a tensor $\mathcal A$ if $\mathcal{A}^{\dagger}$ satisfies $\mathcal{A}*\mathcal{A}^\dagger*\mathcal{A}=\mathcal{A}$ and $\mathcal{A}^\dagger*\mathcal{A}*\mathcal{A}^\dagger=\mathcal{A}^\dagger$.

\begin{definition}[t-SVD]
Given  $\mathcal{A}\in\mathbb{R}^{n_1\times n_2\times n_3}$, the $t$-SVD of $\mathcal{A}$ is given by 
\[\mathcal{A}=\mathcal{U}*\mathcal{S}*\mathcal{V}^\top,
\]
where $\mathcal{U}\in\mathbb{R}^{n_1\times m\times n_3}$ and $\mathcal{V}\in\mathbb{R}^{n_2\times m\times n_3}$ are orthogonal tensors,   $\mathcal S\in\mathbb{R}^{m\times m\times n_3}$ is a f-diagonal tensor (each frontal slice is a diagonal matrix), and $m=\min\{n_1,n_2\}$.
\end{definition}


\begin{definition}[Tensor multi rank and tubal rank]
The tensor multi rank of a 3-mode tensor $\mathcal{A}\in\mathbb{R}^{n_1\times n_2\times n_3}$ is a vector $\h r\in\mathbb{R}^{n_3\times 1}$ with its $i$-th component equal to the rank of the $i$-th frontal slice of $\hat{\mathcal{A}}$. The tensor tubal rank of $\mathcal{A}$ is defined to be $r=\|\h{r}\|_{\infty}$.
\end{definition}


\begin{definition}[t-CUR]
For 
$\mathcal{A}\in\mathbb{R}^{n_1\times n_2\times n_3}$, the t-CUR decomposition of $\mathcal{A}$ is given by
$\mathcal{C}*\mathcal{U}^\dagger*\mathcal{R}$, 
where $\mathcal{C}=[\mathcal{A}]_{:,J,:}$, $\mathcal{R}=[\mathcal{A}]_{I,:,:}$, $\mathcal{U}=[\mathcal{A}]_{I,J,:}$  with  $I\subseteq[n_1]$ and $J\subseteq[n_2]$. 
\end{definition}
When the multi ranks of $\mathcal{U}$ and $\mathcal{A}$ are the same, Chen et al.~\cite{chen2022tensor} proved that the t-CUR representation of $\mathcal{A}$ is exact, i.e.,  $\mathcal{A}=\mathcal{C}*\mathcal{U}^\dagger*\mathcal{R}$. 

\medskip
In this work, we focus on a specific sampling strategy, referred to as tubal sampling~\cite{zhang2016exact}, under which one can  randomly select  tensor tubes to sample. Specifically for a  tensor of dimension ${n_1\times n_2\times n_3}$, one defines a  sampling operator $\mathcal{P}_{\Omega}$   by
\begin{equation}\label{eqn:sampling}
    [\mathcal{P}_{\Omega}(\mathcal{X})]_{i,j,k}:=\begin{cases}
[\mathcal{X}]_{i,j,k},& \text{if}~(i,j,k)\in\Omega\\
0,& \text{Otherwise},
\end{cases}
\end{equation} 
that corresponds to an index set
\begin{equation}\label{eq:tubalsampling}
  \Omega=\{(i,j,k):(i,j)\in  \Phi\subseteq [n_1]\times[n_2], \forall k\in[n_3]\}.  
\end{equation}
In tubal sampling \eqref{eq:tubalsampling}, each  tubal is either sampled entirely or not sampled at all. We will develop two non-convex methods for tensor completion under tubal sampling: one is by imposing regularizations (Section~\ref{sec:ADMM}) and the other is via t-CUR (Section~\ref{sec:CUR}).

\section{Tensor Low-rank regularization}\label{sec:ADMM}

Recovering a (complete) tensor $\mathcal X$ from its partial observations $\mathcal Y=\mathcal{P}_\Omega(\mathcal X)$ is a highly ill-posed problem. We are interested in finding a tensor with  small tubal rank by imposing a proper regularization, denoted by $h(\cdot)$. We consider a general model for low-rank tensor completion
\begin{equation}
 \min_{\mathcal X\in \mathbb{R}^{n_1\times n_2\times n_3}} h(\mathcal X) \quad s.t. \quad \mathcal{Y}=\mathcal{P}_\Omega(\mathcal X). \label{eq:4_Ch02}
\end{equation}
By introducing an auxiliary variable $\mathcal Z$, we adopt the alternating direction method of multipliers (ADMM) \cite{boyd2010distributed} that iterates as follows,
\begin{align*}
    \mathcal{X}^{(\ell+1)} &= \arg\min_\mathcal{X}\{ ||\mathcal{X} - (\mathcal{Z}^{(\ell)} - \mathcal{B}^{(\ell)})||_\text{F}^2 \ \mbox{s.t.} \ \mathcal{Y}=\mathcal{P}_\Omega(\mathcal X) \}\\ 
    \mathcal{Z}^{(\ell+1)} &= \arg\min_\mathcal{Z}\{\frac{1}{\rho}h(\mathcal Z) + \frac{1}{2}||\mathcal{Z} - (\mathcal{X}^{(\ell+1)} + \mathcal{B}^{(\ell)})||_\text{F}^2\}\\
    \mathcal{B}^{(\ell+1)} &= \mathcal{B}^{(\ell)}+(\mathcal{X}^{(\ell)}-\mathcal{Z}^{(\ell+1)}),
\end{align*}
where $\mathcal B$ is a Lagrangian multiplier to enforce $\mathcal X=\mathcal Z$, $\rho>0$ is a weighting parameter, and $\ell$ counts the iterations. The algorithm alternates between $\mathcal X$ satisfying the data matching constraint and promoting $\mathcal Z$ to be low-rank. The closed-form solution for $\mathcal X^{(\ell+1)}$ is that it takes the values of $\mathcal Y$ on $\Omega$ and of $\mathcal Z^{(\ell)}-\mathcal B^{(\ell)}$ on the complement set of $\Omega$.

A popular choice of  $h(\cdot)$ is the tensor nuclear norm (TNN)~\cite{6909886}, defined by 
\begin{equation}
    ||\mathcal{X}||_{\text{TNN}} = \sum_{j=1}^{n_3} \sum_{i=1}^{m} [\mathcal S]_{i,i,j},
    \label{eqn:TNN}
\end{equation}
where $\mathcal X\in\mathbb{R}^{n_1\times n_2\times n_3}$ has the t-SVD of $\mathcal X=\mathcal U*\mathcal S*\mathcal V^\top$ and $m=\min(n_1,n_2)$. The algorithm that minimizes the TNN via ADMM is referred to as TNN-ADMM \cite{LiuX2020,popa2021improved}. The $\mathcal Z$-subproblem has a closed-form solution, referred to as tensor singular value thresholding \cite{lu2019tensor},
\begin{equation}
    \mathcal Z^{(\ell+1)} = \mathcal U*\mathcal S_{1/\rho}*\mathcal V^\top,
\end{equation}
where  t-SVD of $\mathcal X^{(\ell+1)}+\mathcal B^{(\ell)}$ is given by $\mathcal U*\mathcal S*\mathcal V^\top$ and $\mathcal S_{\mu}=\ifft(\max(\hat{\mathcal{S}}-\mu,0),[~],3).$ 

We propose the tensor $L_1$-$L_2$ (TL12) regularization, 
\begin{equation}
    ||\mathcal{X}||_{\text{TL12}} = \sum_{j=1}^{n_3} \left(\sum_{i=1}^m [\mathcal S]_{i,i,j}- \sqrt{\sum_{i=1}^m [\mathcal S]^2_{i,i,j}} \right),
    \label{eqn:TL12}
\end{equation}
By defining a vector $\h s_j=([\mathcal S]_{i,i,j})_{i=1,\cdots,m}$, the TL12 regularization is equivalent to the difference between the $L_1$ and $L_2$ norms of $\h s_j$, followed by summing over $j=1,\dots, n_3.$ The closed-form solution for the  $\mathcal Z$-subproblem is to replace $\max(\hat{\mathcal S}-\mu,0)$ in defining $\mathcal S_\mu$ by the proximal operator of $L_1$-$L_2$ formulated in \cite{louY18}.

\section{Tensor Completion Based on  CUR Decompositions}\label{sec:CUR}

In this section,  we develop a Tensor Completion method based on   CUR decompositions termed TCCUR. 
We consider the tubal sampling operator $\mathcal{P}_{\Omega}$  defined in \eqref{eqn:sampling} and  the observed data $\mathcal Y=\mathcal{P}_{\Omega}(\mathcal X)$. By the design \eqref{eq:tubalsampling} that each tubal $[\mathcal{X}]_{i,j,:} (\forall (i,j)\in\Phi)$ is completely sampled, we have $\hat{\mathcal{Y}}=\mathcal{P}_{\Omega}(\hat{\mathcal{X}}).$ As a result, we can find an estimate of $\mathcal X$ by completing $\hat{\mathcal{Y}},$ followed by the inverse Fourier transform along the third dimension. Completing the tensor $\hat{\mathcal{Y}}$ reduces to a series of matrix completion problems, independently for each frontal slice of $\hat{\mathcal{Y}}$. 

 Due to tubal sampling \eqref{eq:tubalsampling}, the sampling set is fixed by $\Phi$ for all the frontal slices, and  we denote the sampling operator for matrix completion by $\mathcal P_\Phi$.   
 To complete each frontal slice, we adopt a recently developed matrix completion method termed  iterative CUR completion (ICURC) (see~\cite{cai2022matrix}). Instead of sampling the same row/column indices during the iterations, 
 we randomly generate row and column index sets $I, J$ at each iteration. 
 In other words, we incorporate a resampling strategy into ICURC , hence the name ICURC with resampling (ICURC-R). 
In order to combine the completion results by t-CUR decomposition, all completed matrices only return the completed  row and column submatrices with the same row and column indices $\mathcal{I}$ and $\mathcal{J}$. 

Suppose $Y$ be a matrix of $n_1\times n_2$ that corresponds to any frontal slide of $\hat{\mathcal Y}.$
  We set the initial condition as $X^{(0)}=0.$ Then at every iteration $\ell,$ suppose we have the CUR decomposition of $X^{(\ell)}=C^{(\ell)}(U^{(\ell)})^\dagger R^{(\ell)}.$ 
 The gradient descent update yields
\begin{align*}
    [C^{(\ell+1)}]_{{I}^{c},:} &=[X^{(\ell)}]_{ {I}^c, {J}}+[Y-\mathcal{P}_{\Phi}(X^{(\ell)})]_{ {I}^c, {J}},\\
    [R^{(\ell+1)}]_{:,J^{c}}&=[X^{(\ell)}]_{I,J^c}+[Y-\mathcal{P}_{\Phi}(X^{(\ell)})]_{I,J^c}
\end{align*}
where $I^{c}=[n_1]\setminus I$ and $J^{c}=[n_2]\setminus J$.  The update of $U^{(\ell+1)}$ requires the best rank $r$ approximation, which can be achieved by truncating the largest $r$ singular values in the matrix SVD, denoted by  $\mathcal{H}_r$. In short, we have the formula,
$
    U^{(\ell+1)}=\mathcal{H}_r\left([X^{(\ell)}]_{I,J}+[Y-\mathcal{P}_{\Phi}(X^{(\ell)})]_{I,J}  \right).
$
We stop the iterations when 
\begin{equation*}
    e^{(\ell)}:= \frac{\|[Y-\mathcal{P}_{\Phi}(X^{(\ell)})]_{I,:}\|_\fro+\| [Y-\mathcal{P}_{\Phi}(X^{(\ell)})]_{:,J}\|_\fro}{\|[Y]_{I,:}\|_\fro+\| [Y]_{:,J}\|_\fro} <\varepsilon.
\end{equation*}
for a preset tolerance $\varepsilon>0$. We thus
obtain the row and  column submatrices $[C^{(\ell+1)}]_{\mathcal{I},:}$, $[R^{(\ell+1)}]_{:,\mathcal{J}}$, and $\mathcal{H}_r([X^{(\ell+1)}]_{\mathcal{I},\mathcal{J}})$. The details of ICURC-R are summarized in Algorithm~\ref{ICURC}.

\begin{algorithm}
\caption{Iterative CUR for matrix Completion with Resampling (ICURC-R) } \label{ICURC}
\DontPrintSemicolon
    \KwInput{Observed matrix $Y = \mathcal P_\Phi(X)\in\mathbb{R}^{n_1\times n_2}$; 
  Predefined row and column indices $\mathcal{I},\mathcal{J}$; target rank $r$; error tolerance $\varepsilon;$ maximum iteration number $M$.}

    $X_0=0$, $\ell=0$\\
    \While {$e^{(\ell)}>\varepsilon$ or $\ell\le M$}{Randomly sample row and column indices $I$, $J$ with $|I|=|\mathcal{I}|$ and $|J|=|\mathcal{J}|$.\\
        Set $I^{c}=[n_1]\setminus I$, $J^{c}=[n_2]\setminus J$.\\
        $[R^{(\ell+1)}]_{:,J^{c}}=[X^{(\ell)}]_{I,J^c}+ [Y-\mathcal{P}_{\Phi}(X^{(\ell)})]_{I,J^c}$\\
        $[C^{(\ell+1)}]_{I^{c},:}=[X^{(\ell)}]_{I^c,J}+ [Y-\mathcal{P}_{\Phi}(X^\ell)]_{I^c,J}$\\
        $U^{(\ell+1)}=\mathcal{H}_r\left([X^{(\ell)}]_{I,J}+ [Y-\mathcal{P}_{\Phi}(X^{(\ell)})]_{I,J}  \right)$\\
            $[R^{(\ell+1)}]_{:,J}=U^{(\ell+1)}$\\
            $[C^{(\ell+1)}]_{I,:}=U^{(\ell+1)}$\\
        $X^{(\ell+1)}$ = $C^{(\ell+1)}(U^{(\ell+1)})^{\dagger}R^{(\ell+1)}$\\
        $\ell = \ell + 1$\\
    }

    \KwOutput{ $[X^{(\ell)}]_{:,\mathcal{J}},(\mathcal{H}_r([X^{(\ell)}]_{\mathcal{I},\mathcal{J}}))^\dagger,[X^{(\ell)}]_{\mathcal{I},:}$: estimates of CUR components of $X$.}
\end{algorithm} 

    

For every frontal slice of $[\hat{\mathcal Y}]_{:,:,k}$, 
ICURC-R returns three matrices $[\hat{\mathcal{C}}]_{:,:,k}, [\hat{\mathcal{U}}]_{:,:,k},$ and $[\hat{\mathcal{R}}]_{:,:,k}$. By combining all the matrices as frontal slices, we obtain three corresponding tensors
  $\hat{\mathcal{C}}, \hat{\mathcal{U}}$, and  $\hat{\mathcal{R}}.$ Then we take their inverse Fourier transforms, followed by  t-product, i.e., $\widetilde{\mathcal{X}}= \mathcal{C}\ast \mathcal{U}\ast \mathcal{R}$ as an estimate of $\mathcal{X}$. 

\section{Experiments on completion algorithms}
\label{sec:experiments}

We compare the performance of the two regularizations (TNN and TL12) and one decomposition method (TCCUR) on both synthetic data  and a color image inpainting problem. We generate the sampling set $\Phi\subseteq[n_1]\times[n_2]$ uniformly at random under a preset sampling ratio (without replacements) to define the index set $\Omega$ in \eqref{eq:tubalsampling}. We evaluate the performance by the relative error (RE) and peak signal-to-noise ratio (PSNR), i.e.,
\[\mbox{RE}=\frac{\|\mathcal{X}-\tilde{\mathcal{X}}\|_{\fro}}{\|\mathcal{X}\|_{\fro}} \ \mbox{and} \     \textnormal{PSNR}=10 \log_{10}\left(\frac{n_1n_2n_3\mathcal{X}^2_{\text{max}}}{ \|\tilde{\mathcal{X}}-\mathcal{X}\|^2_\fro}\right),
\]
where $\tilde{\mathcal{X}}\in\mathbb{R}^{n_1\times n_2\times n_3}$ is the recovered tensor and $\mathcal X$ is the ground truth with its maximum absolute value, denoted by $\mathcal{X}_{\text{max}}$.
All   simulations were performed on a laptop with 2.30 GHz Intel(R) Core(TM) i7-11800H processor and 16GB RAM.

\subsection{Synthetic data}
\label{sec:synthetic}

\begin{figure}
\centering
 \includegraphics[width=0.49\linewidth]{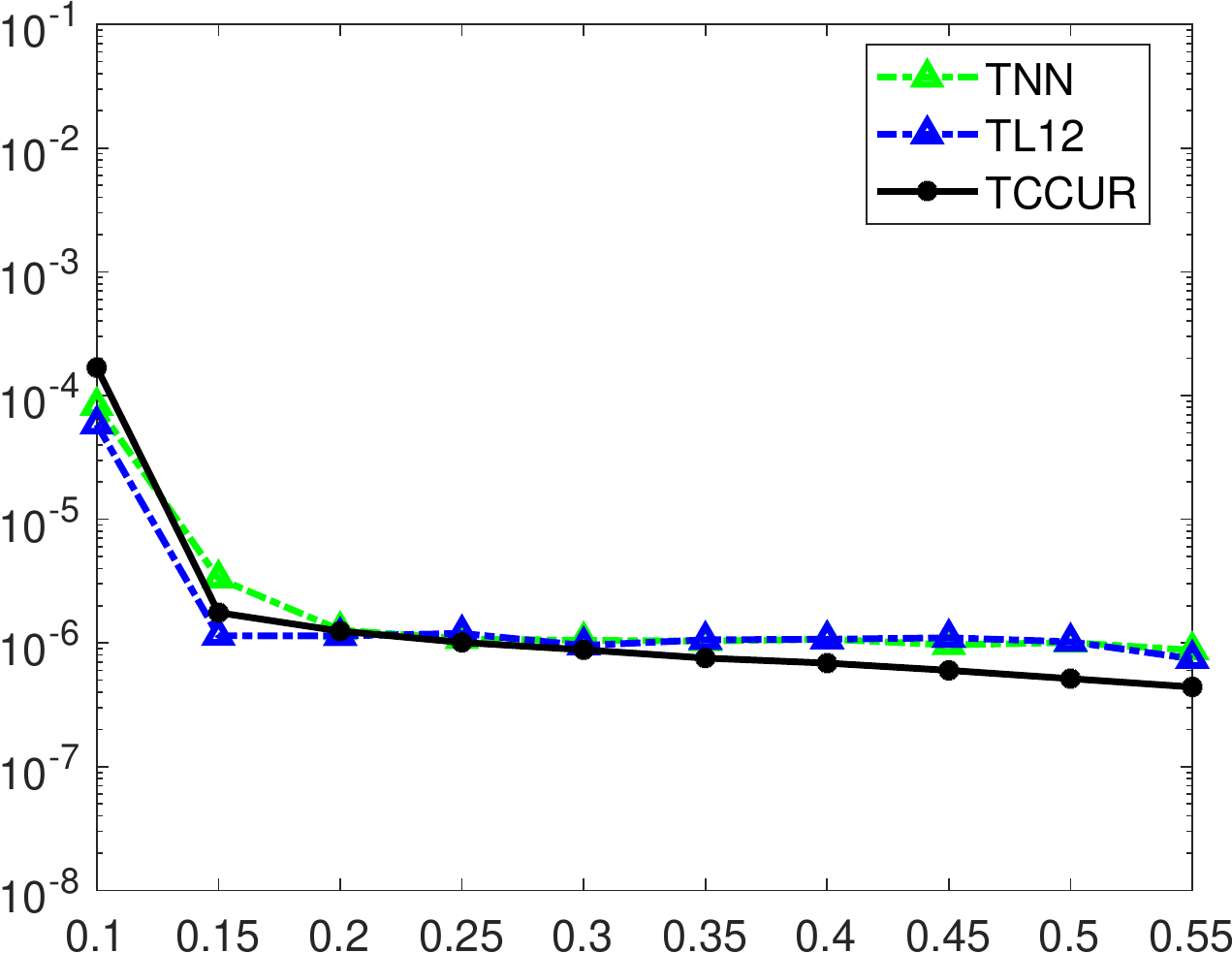}  
 \includegraphics[width=0.49\linewidth]{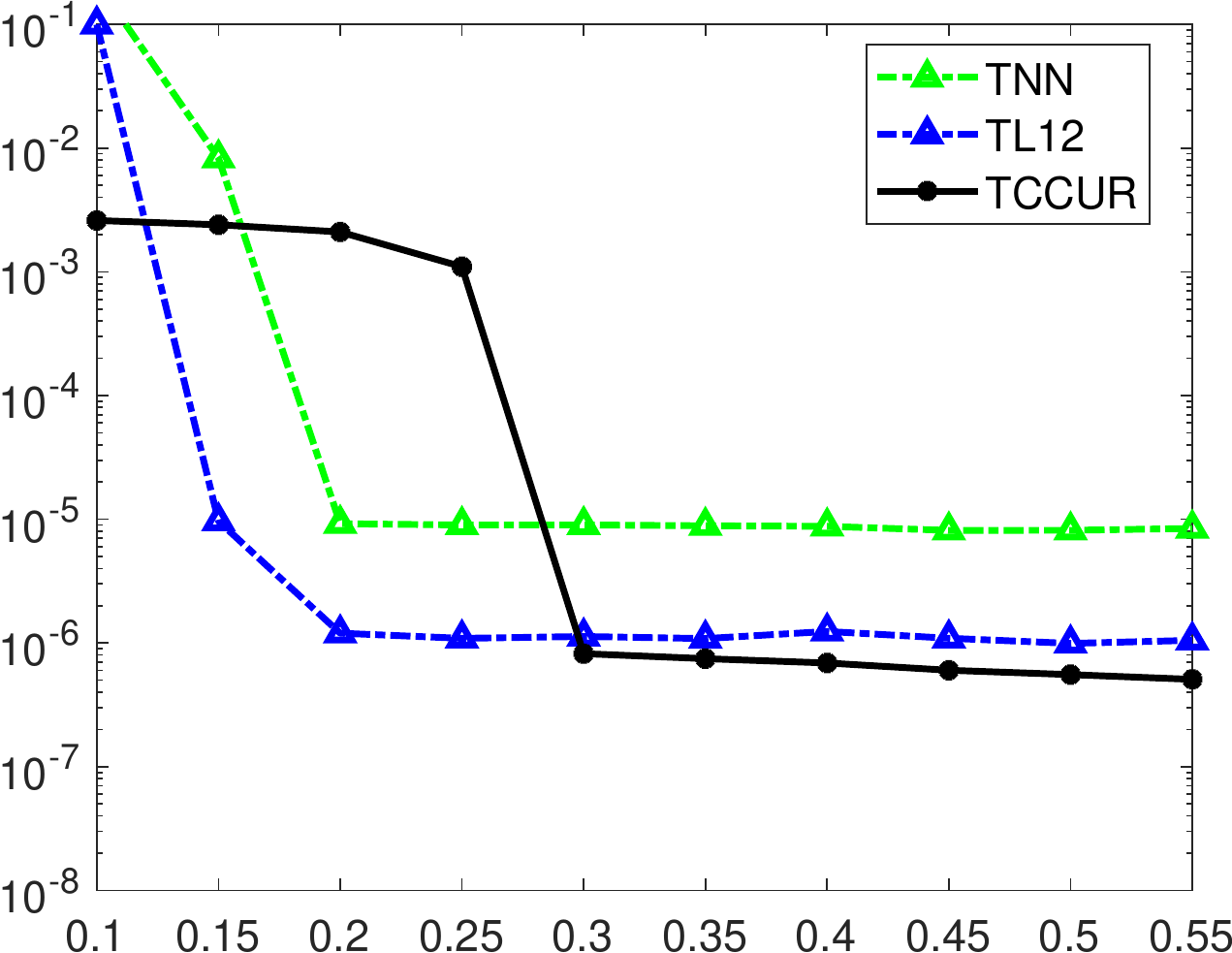}
\caption{
\small REs of completing an underlying tensor of tubal rank 3 (left) and 5 (right) versus SRs. 
}
\label{fig:syndatacompletion}
\end{figure}

\begin{figure}
    \centering
     \includegraphics[width= 0.49\linewidth]{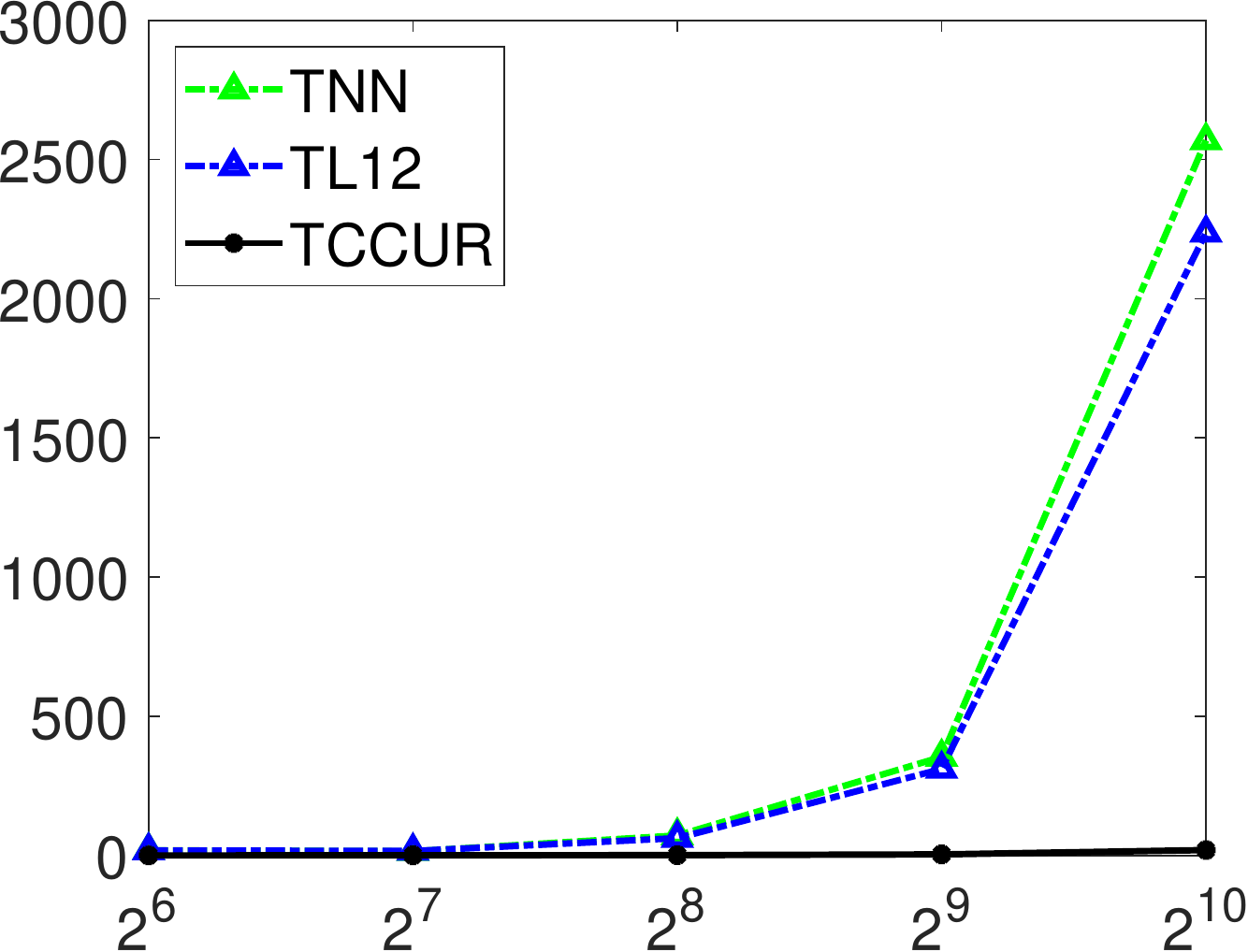}
     \includegraphics[width= 0.49\linewidth]{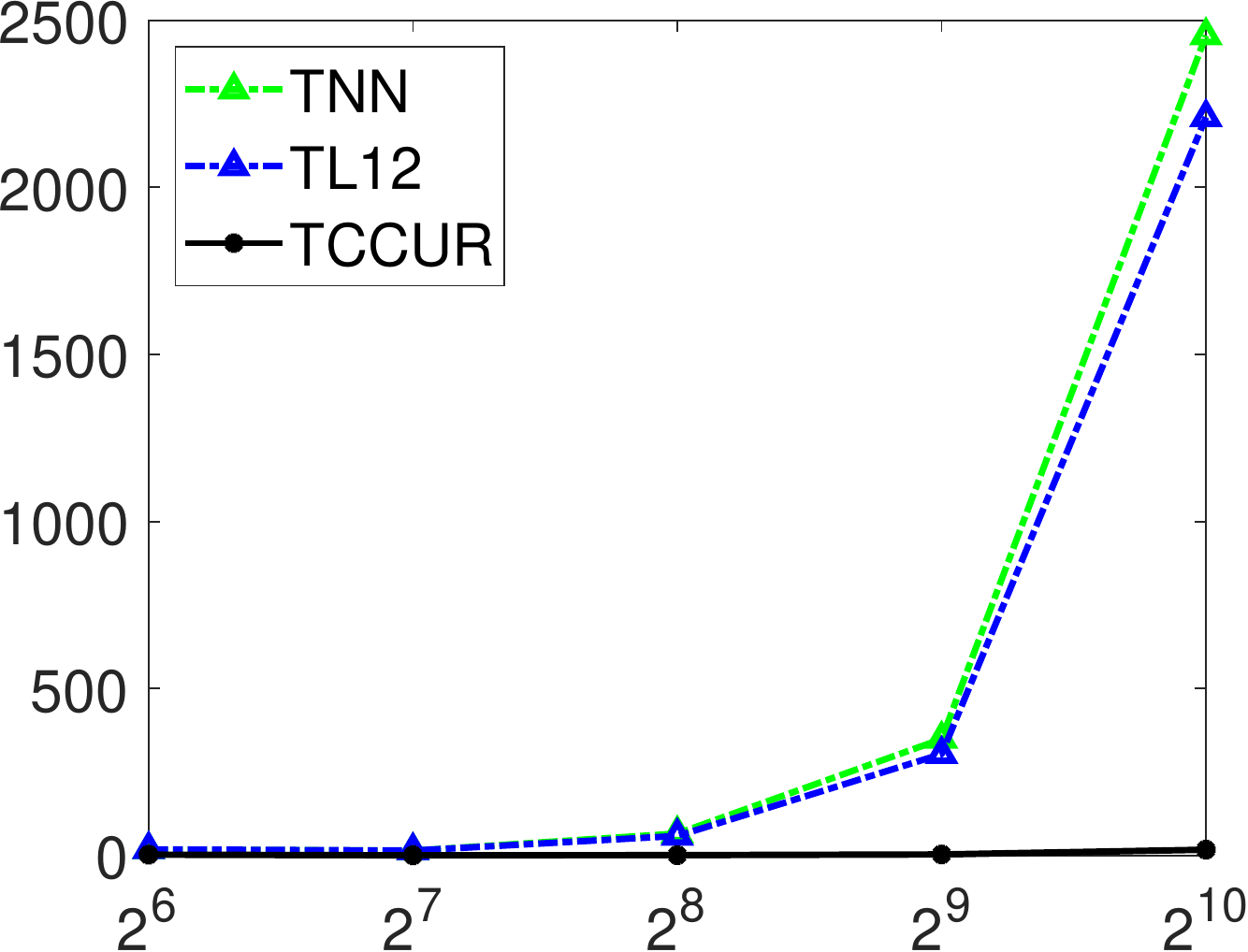}
    \caption{\small Runtime versus the frontal slice dimension  $2^n$  of $2^n\times 2^n\times 32$ tensors with tubal rank 2 (left) and 3 (right).}
    \label{fig:time}
\end{figure}

We generate a tensor $\mathcal{X}\in\mathbb{R}^{256\times 256\times 50}$ with tubal rank $r\in\{3,5\}$. For each preset rank, Fig.~\ref{fig:syndatacompletion} shows the mean of REs over 30 random realizations with respect to sampling ratio (SR), showing that both TL12 and TCCUR achieve comparable and even better performance than  TNN. Moreover, the TL12 method has  the fastest decay of RE for smaller SRs (e.g., $10\%-20\%$).




We also examine the scalability of the algorithms by reporting the runtime with respect to the tensor's dimensions. In particular, we generate a tensor $\mathcal{X}\in\mathbb{R}^{2^n\times 2^n\times 32}$ with tubal rank $2$ and $3$ for $n=6, 7, 8, 9, 10.$ We randomly select $30\%$ tubals and adopt the same stopping condition for all the algorithms, that is, the relative error on the observed portion is less than 
 $10^{-6}$. The computational time
 is reported in Fig.~\ref{fig:time}, illustrating significant advantages in the efficiency of TCCUR over TNN and TL12. In addition, TL12 is comparable in speed compared to TNN, yet gives better completion accuracy. 
\subsection{Application on Image Inpainting}
\label{sec:realdata}

\begin{figure}[h]
\centering
\includegraphics[width=.24\linewidth]
{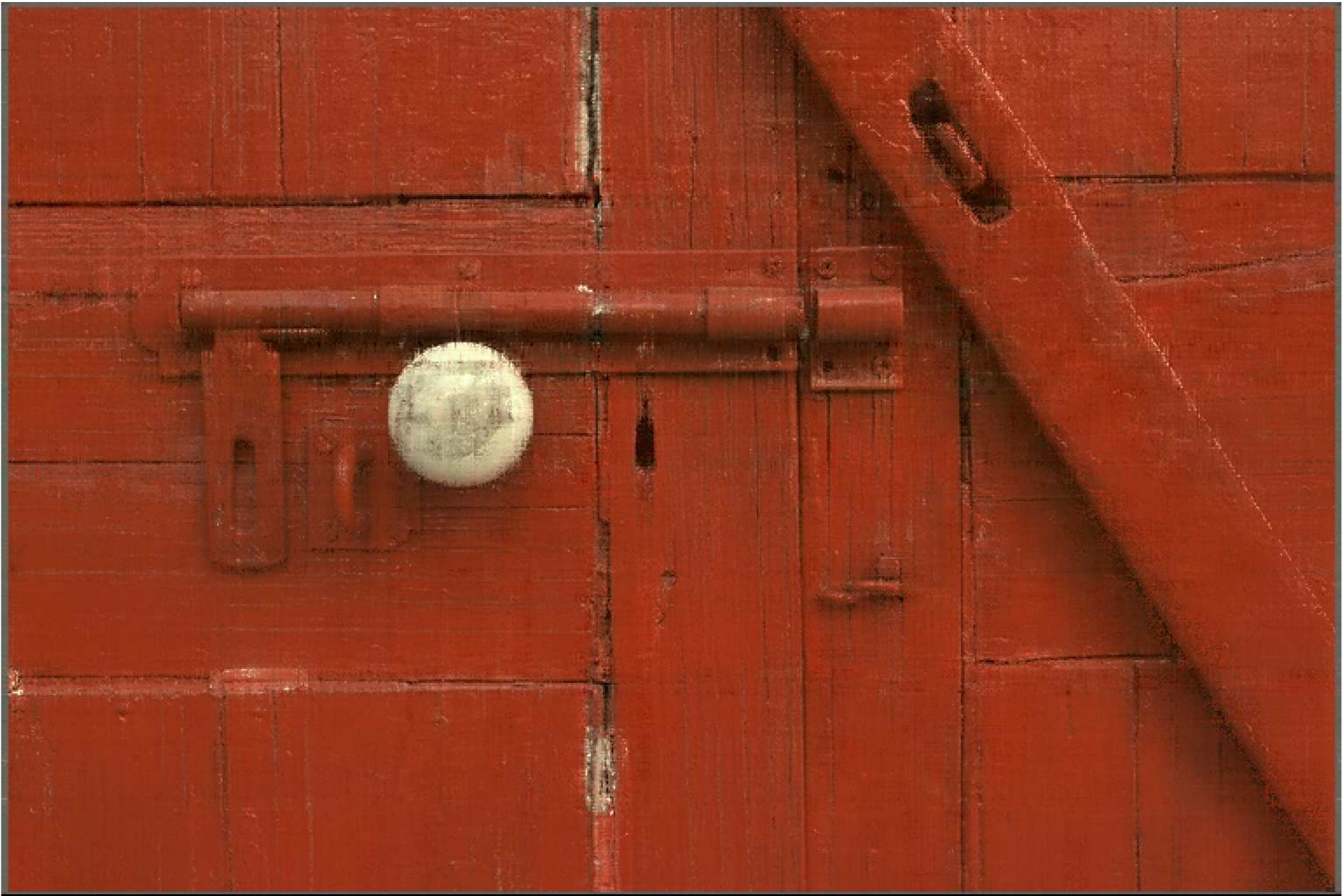}
\includegraphics[width=.24\linewidth]
{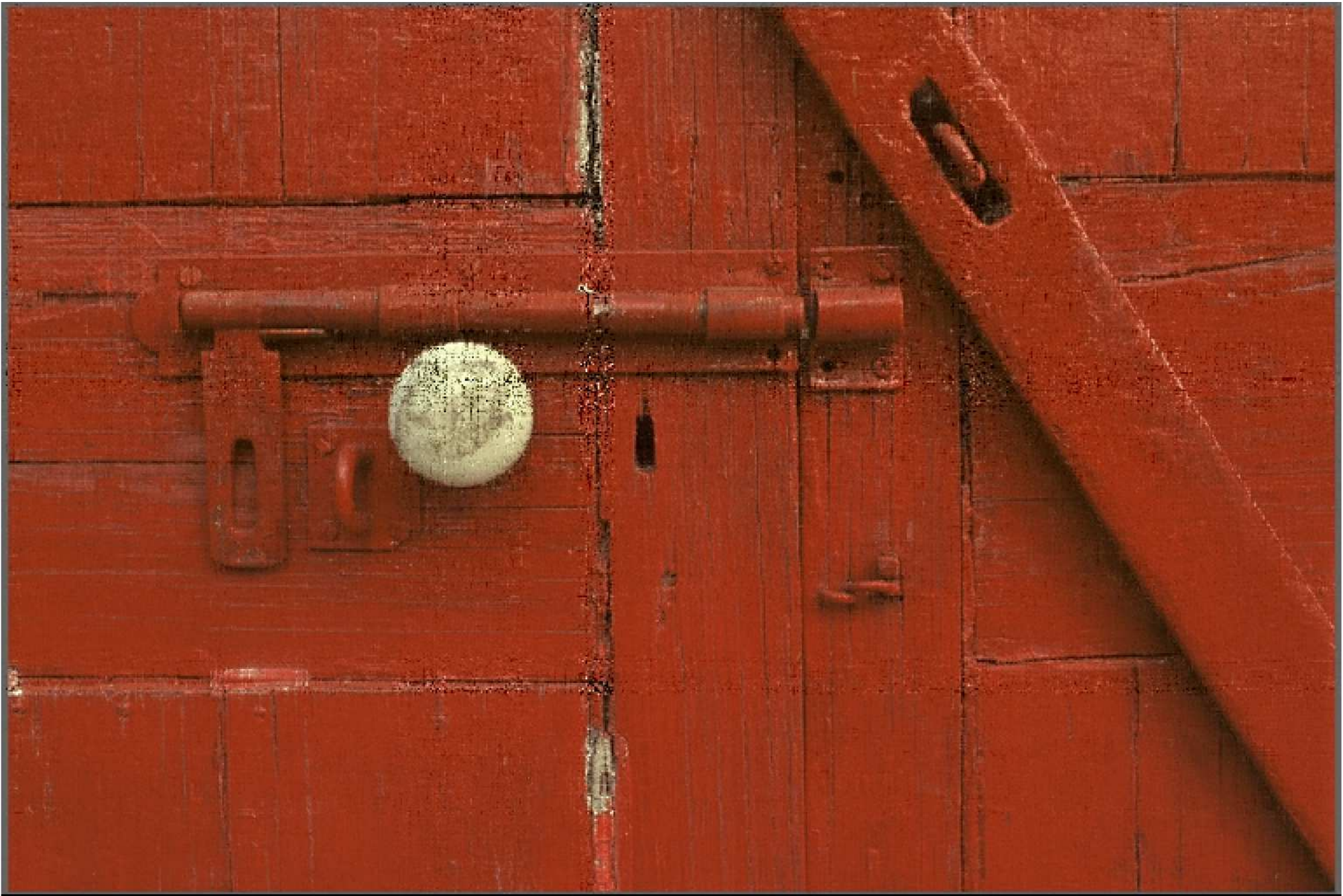}
\includegraphics[width=.24\linewidth]
{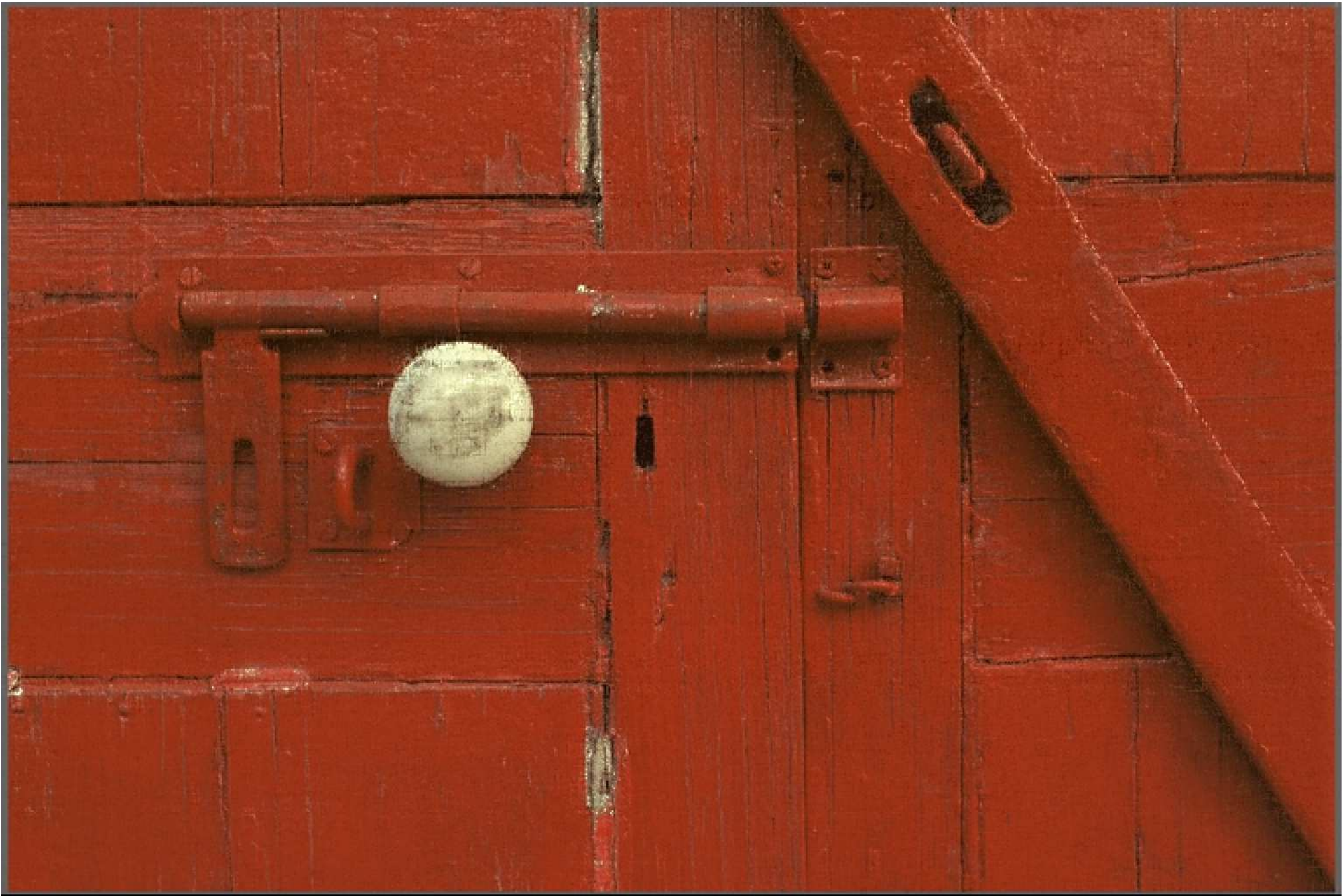}
\includegraphics[width=.24\linewidth]
{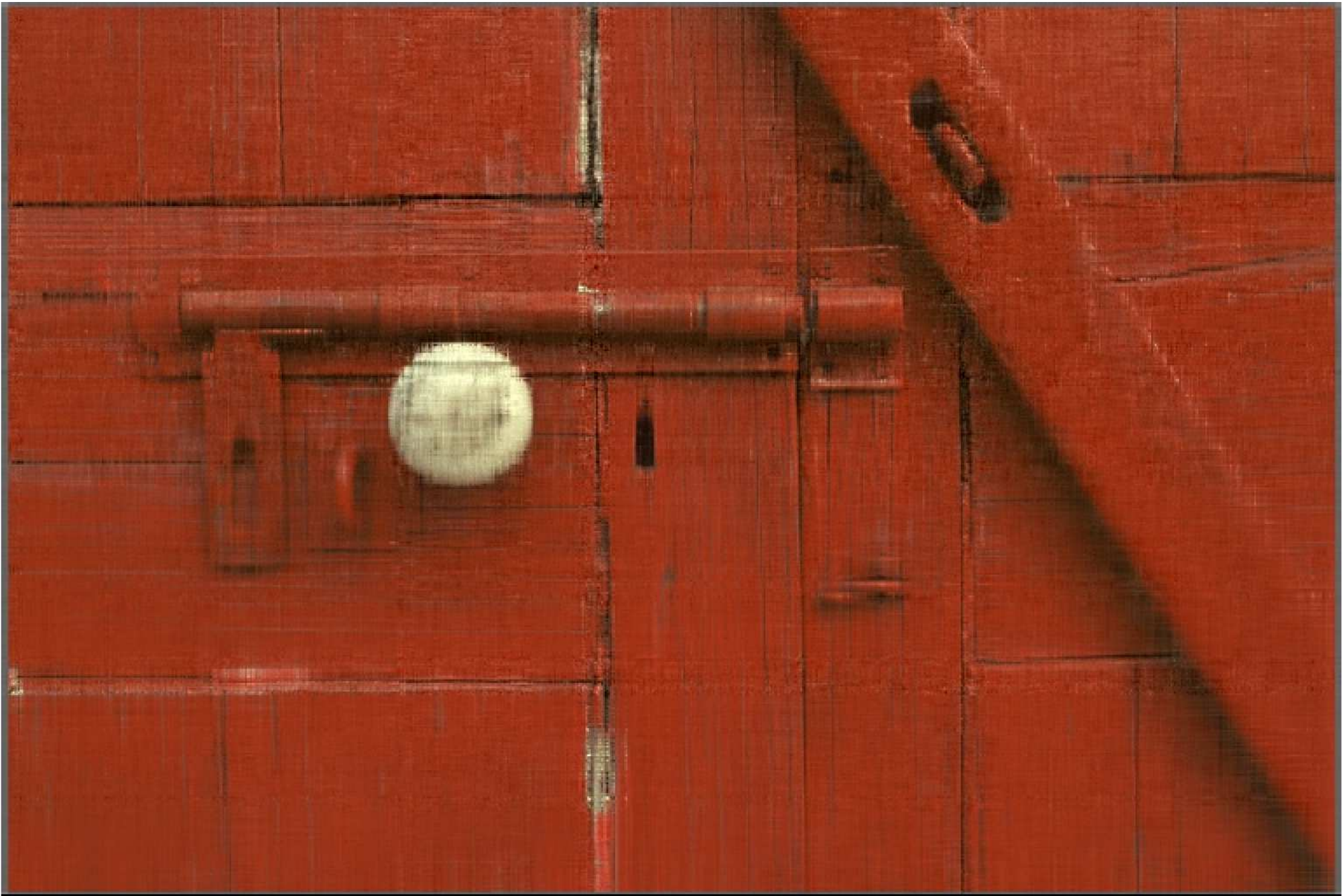}
\\
\includegraphics[width=.24\linewidth
]{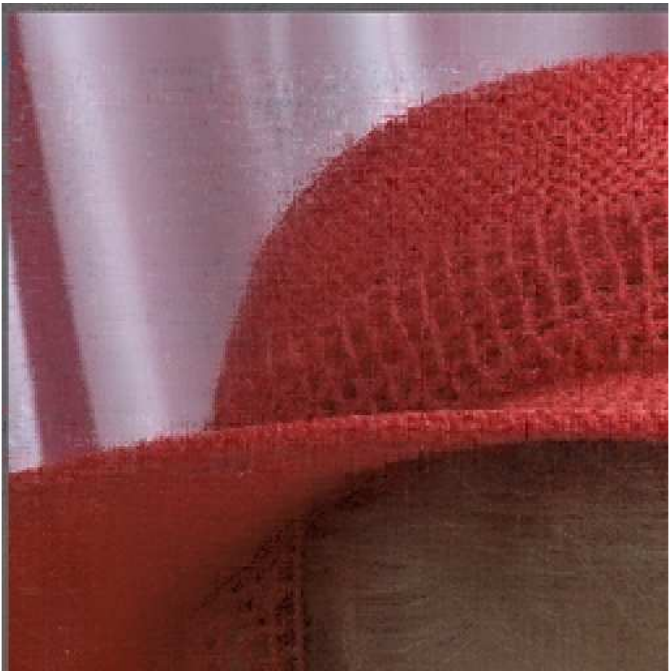}
\includegraphics[width=.24\linewidth 
]{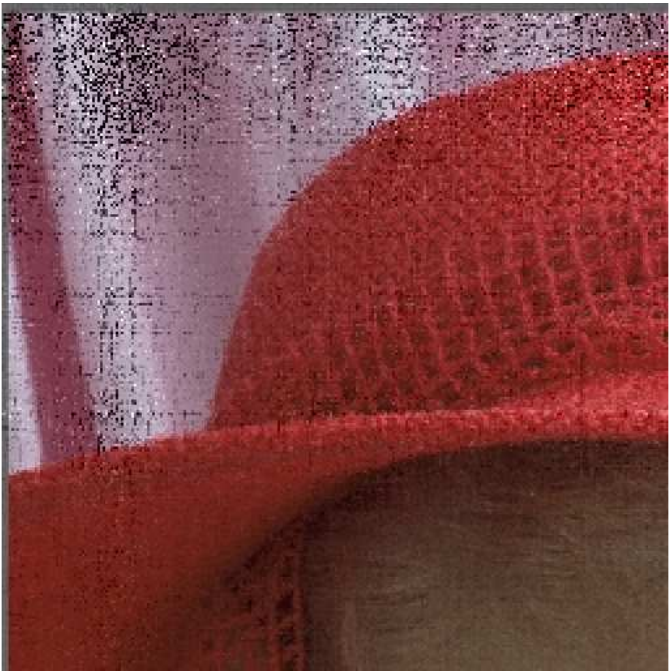}
\includegraphics[width=.24\linewidth 
]{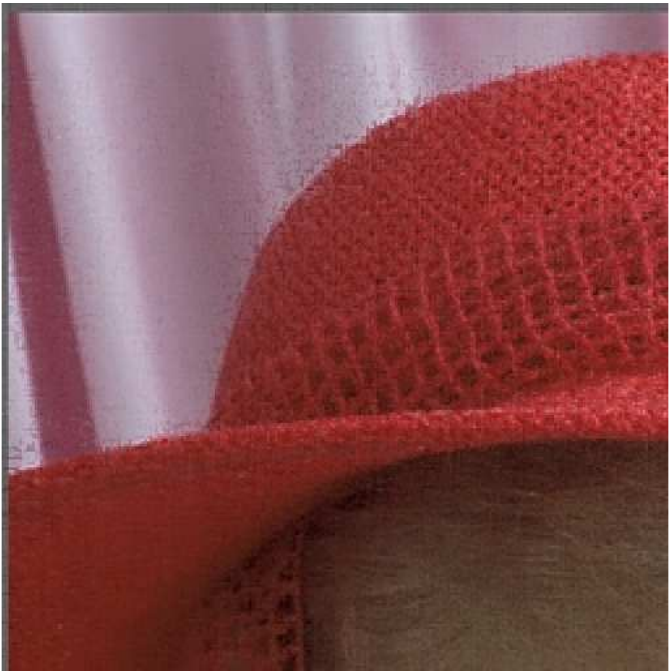}
\includegraphics[width=.24\linewidth 
]{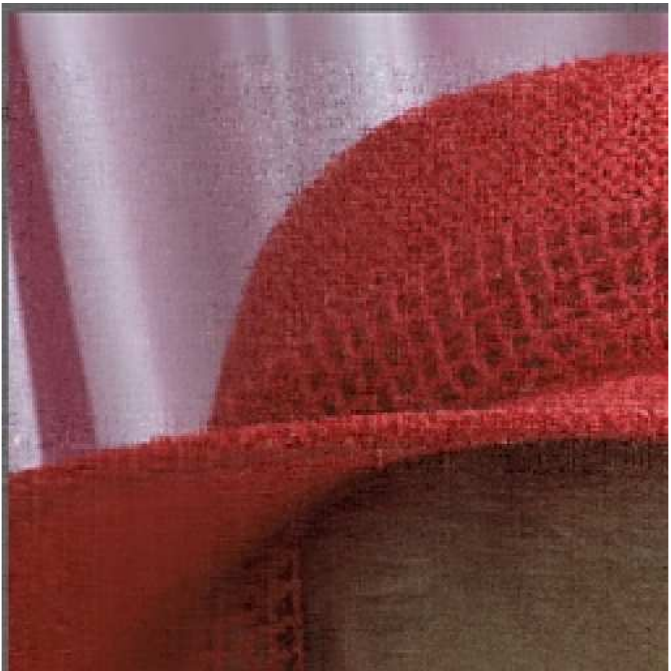}
\\
\includegraphics[width=.24\linewidth
]{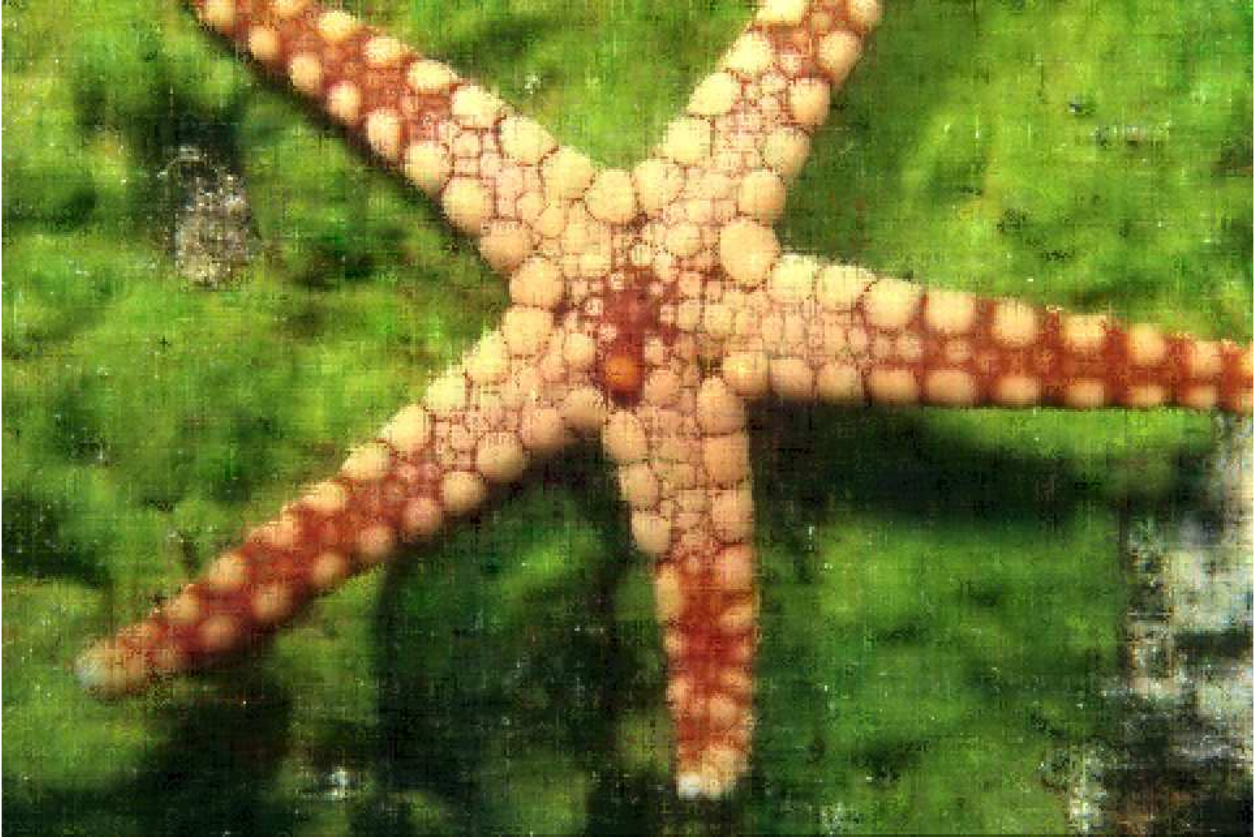}
\includegraphics[width=.24\linewidth 
]{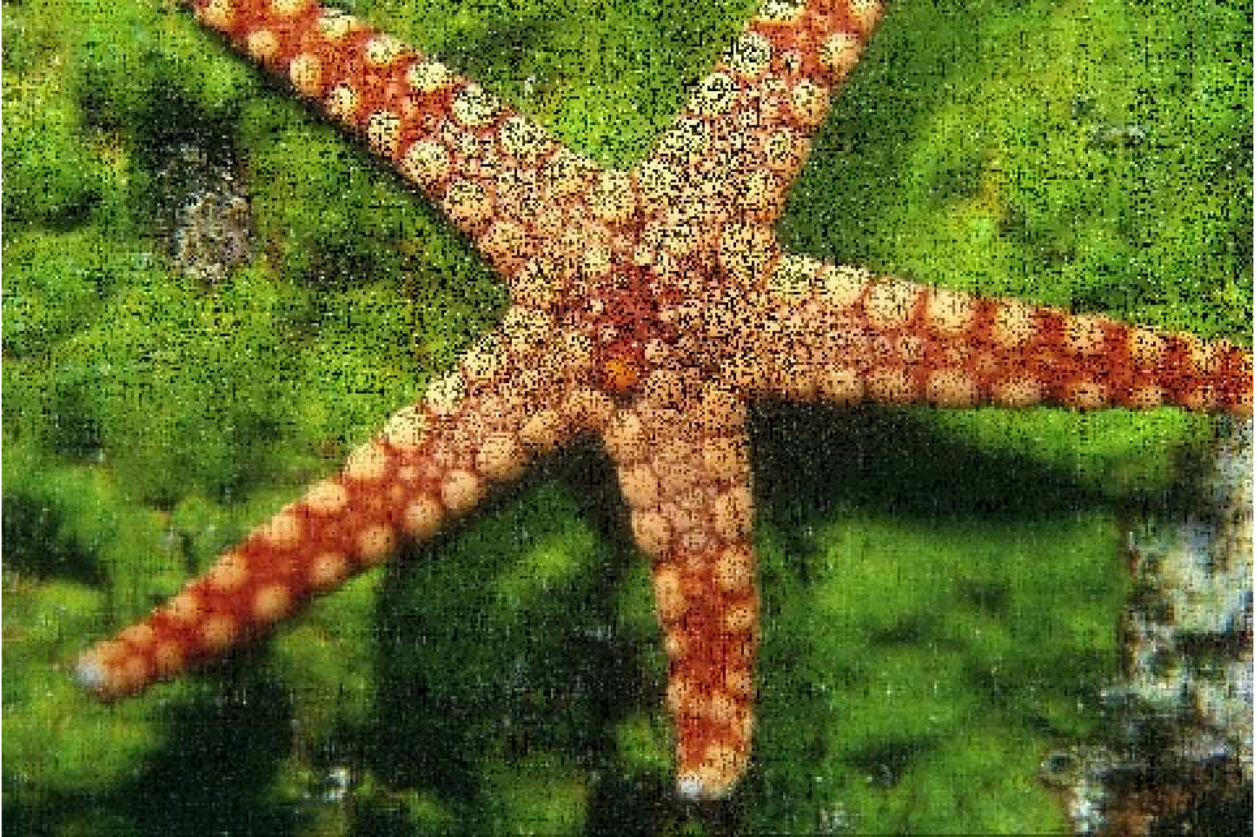}
\includegraphics[width=.24\linewidth 
]{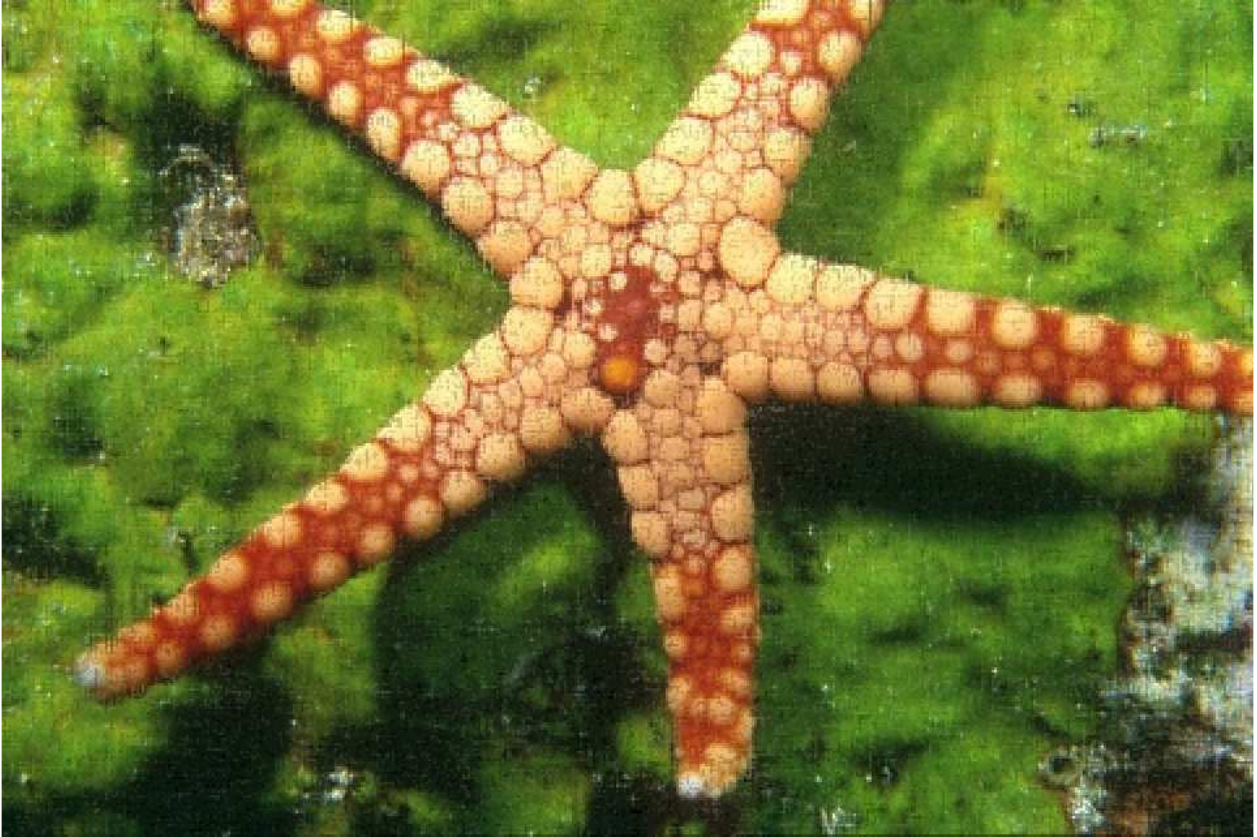}
\includegraphics[width=.24\linewidth 
]{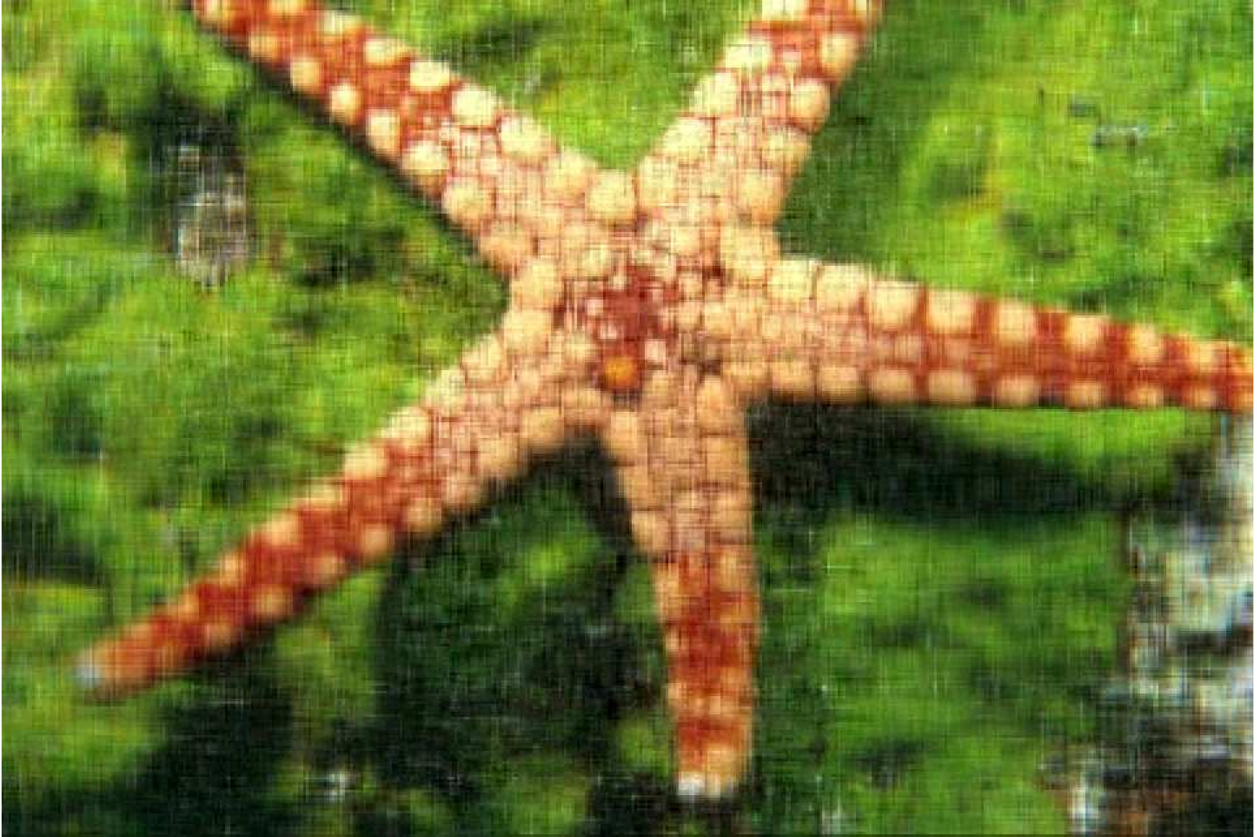}
\begin{minipage}{.24\linewidth} \centering \small TRLRF \end{minipage}
\begin{minipage}{.24\linewidth} \centering \small PSTNN\end{minipage}
\begin{minipage}{.24\linewidth} \centering \small  TL12\end{minipage}
\begin{minipage}{.24\linewidth} \centering \small  TCCUR\end{minipage}
\caption{Visual comparison of color image inpainting results with two state-of-the art methods. From top to bottom, images are labeled as ``Door,'' `` Hat,'' and ``Starfish,'' all are taken from the PSTNN paper.}\label{fig:img_inpainting}
\end{figure}

 We investigate a real application of image inpainting on three color images used in \cite{Jiang_2020}. 
 We compare the proposed methods to two state-of-the-art methods named TRLRF \cite{yuan2019tensor} and PSTNN \cite{Jiang_2020}. 
 We randomly sample   $50\%$  tubals and compare image recovery results obtained by TRLRF, PSTNN, TL12, and TCCUR. 
 Table~\ref{tab:realdata} reports the quantitative measures of inpainting performance in terms of PSNR and computation time.  TCCUR is significantly faster than other methods, and TL12 yields the best results in all test cases both visually and in terms of SNR, though slower than other methods.
Fig.~\ref{fig:img_inpainting} shows the reconstruction results; PSTNN clearly does not achieve, while TRLRF and TCCUR produce more severe artifacts near the rim of the hat, compared to TL12.

\begin{table}[t]
 \label{tab:realdata}
\centering
\begin{adjustbox}{width=\columnwidth,center}
\begin{tabular}{ c|c c|c c|c c } 

\toprule

~ &  \multicolumn{2}{c|}{Door}  & \multicolumn{2}{c|}{Hat} & \multicolumn{2}{c}{Starfish} \cr

~                    & PSNR &  Time                & PSNR &  Time    & PSNR &  Time        \cr

\midrule

TRLRF   &  30.01    & 5.02s    &  26.55  & 4.54s & 23.93 & 13.85s\cr

PSTNN    &  28.13   &  13.58s  & 19.38    &  2.34s & 16.08 & 4.68s \cr

TL12    &  \textbf{31.13}   & 63.60s   & \textbf{27.12} &  8.59s & \textbf{25.94} & 22.08s \cr

TCCUR    &  28.27   & \textbf{4.96s}   & 26.37    & \textbf{1.22s}& 24.14 & \textbf{4.27s} \cr

\bottomrule

\end{tabular}
\end{adjustbox}
\caption{Quantitative comparison of image inpainting from $50\%$ tubal sampling ratio.}
\label{tab:realdata}
\end{table}

\section{Conclusion and discussion}
This paper proposed two novel non-convex tensor completion methods, namely TL12 and TCCUR. Simulation results   demonstrated the trade-off between accuracy and computational costs by using the two methods; the regularization-based method (TL12) achieves high accuracy in tensor completion but at a cost
of high computational complexity, while the decomposition method (TCCUR) is eﬀicient, but its usage is limited to tubal
sampling. Additionally, we considered a real application of color image inpainting and showed the proposed methods outperform the state-of-the-art methods. 


\newpage



\small
\bibliographystyle{IEEEbib}
\bibliography{strings,refs,sparse}

\end{document}